\documentclass[conference]{IEEEtran}
\IEEEoverridecommandlockouts
\usepackage{cite}
\usepackage{amsmath,amssymb,amsfonts}
\usepackage{algorithmic}
\usepackage{algorithm}
\usepackage{graphicx}
\usepackage{textcomp}
\usepackage{xcolor}
\usepackage{tikz}
\usepackage{booktabs}
\usepackage{tabularx}
\usepackage{multirow}
\usepackage{amssymb}
\usepackage{booktabs}
\usepackage{tabularx}
\usepackage{amssymb}
\usepackage{url}

\newcommand{\cmark}{\checkmark}
\newcommand{\xmark}{\texttimes}
\usetikzlibrary{shapes,arrows,positioning}
\def\BibTeX{{\rm B\kern-.05em{\sc i\kern-.025em b}\kern-.08em
    T\kern-.1667em\lower.7ex\hbox{E}\kern-.125emX}}
\begin{document}

\title{Polypersona: Persona-Grounded LLM for Synthetic Survey Responses}

\author{
\IEEEauthorblockN{
Tejaswani Dash\textsuperscript{1},
Dinesh Karri\textsuperscript{1},
Anudeep Vurity\textsuperscript{1},
Gautam Datla\textsuperscript{2},
Tazeem Ahmad\textsuperscript{3},
Saima Rafi\textsuperscript{4},
Rohith Tangudu\textsuperscript{1}
}

\IEEEauthorblockA{\textsuperscript{1}George Mason University, Virginia, USA; 
Emails: tdash@gmu.edu, dkarri@gmu.edu, avurity@gmu.edu, rtangudu@gmu.edu}

\IEEEauthorblockA{\textsuperscript{2}New Jersey Institute of Technology (NJIT), New Jersey, USA; 
Email: tejagvd6@njit.edu}

\IEEEauthorblockA{\textsuperscript{3}University of Southern Queensland, Queensland, Australia; 
Email: Tazeem.Ahmad@unisq.edu.au}

\IEEEauthorblockA{\textsuperscript{4} Edinburgh Napier University, Edinburgh, Scotland; Email: S.rafi@napier.ac.uk}
}

\maketitle


\begin{abstract}
This paper introduces \textit{PolyPersona}, a generative framework that instantiates a persona-conditioned language model to synthesize realistic survey responses across multiple domains. The framework instruction-tunes compact chat models using parameter-efficient LoRA adapters and 4-bit quantization under a resource-adaptive training setup. A dialogue-formatted data pipeline that preserves persona cues to maintain consistent behavioral alignment across responses\footnote{The implementation and evaluation code are available in an anonymized repository at \url{https://anonymous.4open.science/r/Polypersona-1D70/}}. The resulting dataset comprises 3,568 responses spanning ten domains and 433 unique personas, enabling controlled instruction-tuning and systematic multi-domain evaluation. A multi-metric evaluation stack integrates BLEU, ROUGE, and BERTScore with survey-specific metrics capturing structural, stylistic, and sentiment consistency. Results show that small models such as TinyLlama 1.1B and Phi-2 achieve performance on par with larger 7B–8B baselines (highest BLEU 0.090, ROUGE-1 0.429), highlighting the efficiency of persona-conditioned fine-tuning on compact architectures. Persona-grounded small models prove reliable and efficient for generating synthetic survey data, with open protocols ensuring bias monitoring and reproducibility. 
\end{abstract}


\section{Introduction} \label{intro}

Modern survey research is increasingly strained by rising costs and falling participation \cite{sur}. Traditional probability samples are expensive to recruit and often suffer from missing data due to nonresponse or attrition, which undermines inference and forces reliance on convenience samples~\cite{holtdirk2025,couper2017}. At the same time, response rates continue to decline and coverage gaps widen as certain demographic groups become harder to reach~\cite{bisbee2024}. These issues worsen selection biases and limit diversity, as researchers often lack representation for subpopulations or sensitive topics, hindering balanced insights. In reality, most modern surveys struggle with making accurate conclusions, are becoming more expensive, and have fewer participants, so there is no real replacement for high-quality data directly from respondents~\cite{bisbee2024,holtdirk2025}. Existing computational solutions attempt to alleviate these issues but have clear limitations. Early approaches, such as rule-based simulations or statistical imputation models, fail to capture the multidimensional interplay of attitudes, beliefs, and demographics in human responses. Recent advances in large language models (LLMs) have enabled the simulation of persona-based synthetic respondents, capable of producing fluent, contextually appropriate answers. LLMs have been explored for fine-tuned imputation of missing survey answers, often matching traditional tabular methods in accuracy when data are missing at random~\cite{holtdirk2025}. 
The promise of persona conditioning is to simulate realistic feedback by constraining model behavior with thousands of socioeconomic backstories~\cite{bisbee2024}. 
However, LLMs are not a universal solution as they can assist with question clarity or cognitive load reduction, but they cannot yet correct for sampling and non-response bias~\cite{jansen2023,ahmad2024mlops}. 
Many existing LLM-based generators rely on ad-hoc prompting or lightweight fine-tuning without persona constraints, leading to inconsistent demographic and psychographic grounding~\cite{li2025,bisbee2024,holtdirk2025}. Without structured persona conditioning, these models often drift from their intended profiles, producing biased or repetitive outputs.

Recent evaluations highlight additional limitations of synthetic LLM respondents. Controlled comparisons show that chat-based models (e.g., ChatGPT) reproduce general opinion trends but fail to capture genuine population variance~\cite{bisbee2024,millman2025}. 
Synthetic personas frequently regress toward the mean, yielding homogeneous and overly agreeable responses. 
Bisbee et al.~\cite{bisbee2024} found that GPT-based respondents exhibited reduced variance and systematically different regression estimates compared to real survey data. 
Similarly, Shrestha et al.~\cite{shrestha2025} reported that GPT-4 personas skewed toward more progressive or socially desirable responses, diverging from actual population distributions, particularly for non-WEIRD demographic groups (i.e., populations that are not \textbf{W}estern, \textbf{E}ducated, \textbf{I}ndustrialized, \textbf{R}ich, or \textbf{D}emocratic). Further, small prompt modifications or model updates can shift output distributions over time, compromising reproducibility~\cite{millman2025}. 
These findings highlight persistent reliability and representativeness challenges in current synthetic survey pipelines.

To address these gaps, we propose \textbf{polypersona}, a generative framework that explicitly embeds demographic and psychographic consistency into LLM-driven survey simulations. 
Polypersona instantiates LLMs with grounded persona profiles, ensuring that each maintains coherent behavioral traits and opinions across multiple question types and topics. 
The system leverages structured prompts, dialogue-formatted data, and (optionally) instruction-tuned adapters to enforce within-persona alignment. 
We evaluate the approach using quantitative and qualitative criteria: (a) response quality and plausibility, (b) diversity and demographic representativeness, and (c) persona faithfulness across question modalities. 
Our implementation is released as an open-source toolkit to facilitate reproducible synthetic survey generation in academic and commercial settings.

\textbf{Our key contributions are as follows.}
\begin{enumerate}
    \item \textbf{Persona-based generation methodology:} A systematic procedure for constructing an LLM-based survey model with fixed demographic and psychographic attributes that remain consistent across diverse question types.
    \item \textbf{Evaluation framework:} Quantitative and qualitative metrics for assessing data fidelity, diversity, and persona coherence, enabling rigorous benchmarking against real-world survey datasets.
    \item \textbf{Open-source implementation:} A flexible and accessible toolkit for survey researchers to simulate respondents, pilot questionnaires, and perform controlled sensitivity analysis.
\end{enumerate}

While polypersona is designed to augment human data collection, it is not a replacement for real respondents. As recent work emphasizes,``there is simply no replacement for fully permissioned first-party data from real humans''~\cite{millman2025}. 
Instead, polypersona serves as a complementary capability for pretesting instruments, exploring demographic response patterns, and supporting large-scale synthetic experimentation where traditional recruitment is costly or impractical.

This study explores three research questions:
\begin{itemize}
    \item \textbf{RQ1:} To what extent can LLMs maintain assigned persona characteristics across question modalities and domains?
    \item \textbf{RQ2:} How effectively do persona-based LLMs ensembles generate diverse, representative response distributions compared to human samples?
    \item \textbf{RQ3:} What are the practical boundaries and limitations of persona-infused LLMs in academic and applied survey contexts?
\end{itemize}

By answering these questions, this work contributes to the growing literature on LLM-based synthetic data generation and provides empirical guidance for incorporating persona-conditioned language models within mixed-methods research workflows.

\section{Related Work}

\subsection{Synthetic Data Generation}
Synthetic data generation has emerged as a pivotal solution for addressing data scarcity, privacy concerns, and the need for robust AI training across diverse domains~\cite{abufadda2021,ahmad2024mlops}. Early innovations in survey methodology, such as address-based sampling and mixed-mode data collection, improved representativeness but continued to struggle with selection bias~\cite{couper2017}. Machine learning approaches have leveraged for data synthesis and classification tasks in healthcare, finance, and biometrics, though these methods often fail to capture the multidimensional complexity of human behavior~\cite{lehtonen2025,vurity1, vurity2023new,vurity2024interpreting,abdullah2025hypergcn,faye2024k}.

The advent of large language models (LLMs) has enabled more sophisticated, domain-specific synthetic data generation through techniques like prompt engineering and parameter-efficient fine-tuning~\cite{guo2024}. However, concerns about hallucination, representational bias, and ethical implications persist. S-BERT-based pipelines have demonstrated promising replication of human factors in survey responses by leveraging semantic similarity across items, though limitations remain in handling negations and scaling across diverse survey types~\cite{lehtonen2025}. These gaps underscore the need for approaches that better capture the nuanced relationships between respondent characteristics and response patterns.

Recent studies continue to highlight \textbf{LLM-based synthetic data generation} as a promising solution to survey data scarcity and bias. Argyle\textit{ et al.} demonstrate that carefully persona-conditioned language models can closely emulate real survey populations by conditioning GPT-3 on thousands of socio-demographic backstories, producing nuanced response distributions correlated with human subgroups~\cite{argyle2023out}. This supports the notion that LLMs, when grounded in rich profiles, can act as effective proxies for specific human subpopulations. However, researchers urge caution regarding the reliability and interpretation of such synthetic outputs. Madden in ~\cite{madden2025synthetic} argues that while LLMs are increasingly used to augment survey responses or power multi-agent simulations, their outputs should not be misinterpreted as true population samples. These contributions extend the literature on synthetic data generation beyond earlier GAN-based and imputation approaches, reinforcing the need for robust, demographically grounded generators and ethical safeguards in LLM-synthesized survey data.

\subsection{LLMs for Text Generation}
The rise of LLMs has significantly advanced natural language processing tasks, including synthetic data generation. Modern frameworks such as \textit{DataDreamer} provide unified approaches to address scalability and reproducibility challenges in LLM-driven synthetic data generation and fine-tuning workflows~\cite{patel2024,luo2025large,adnan}. Similarly, \textit{SynthIE} leverages task asymmetry to produce high-quality synthetic datasets~\cite{josifoski2023}, while \textit{SynDG} generates grounded dialogues through structured dialogue flows and filtering strategies to enhance coherence and quality~\cite{bao2023}. Together, these systems represent an evolution toward more systematic and reproducible synthetic data generation pipelines.

In parallel, advances in \textbf{parameter-efficient fine-tuning (PEFT)} have enabled effective customization of LLMs at scale. Methods such as \textit{LoRA}~\cite{hu2021lora} and \textit{QLoRA}~\cite{dettmers2023qlora} significantly reduce computational cost while maintaining performance. LoRA introduces low-rank adaptations to attention layers, achieving near full fine-tuning performance with only 1–2\% of trainable weights, while QLoRA extends this by incorporating 4-bit quantization to fine-tune models with up to 65B parameters on modest hardware. These techniques demonstrate that smaller, efficiently tuned models can rival or exceed larger ones on domain-specific tasks, as observed in our polypersona experiments.

In text classification domains, research has shown that low-subjectivity tasks benefit more from synthetic data generation methods~\cite{li2023}. CodecLM applies instruction tuning with metadata encoding and iterative refinement to produce diverse, high-quality synthetic instructions~\cite{long2024}. A comprehensive survey by Long \textit{et al.} in \cite{long2024} underscores the need for unified methodologies in curating, generating, and evaluating synthetic data across NLP workflows, identifying persistent gaps such as biases in LLM outputs, coherence issues in dialogue generation, and challenges in domain-specific adaptation.
Concurrently, \textbf{prompt engineering} has emerged as a vital complement to fine-tuning. Chen \textit{et al.}~\cite{chen2024prompt} provide a comprehensive review of prompting strategies, emphasizing few-shot and chain-of-thought approaches for enhancing reasoning and task accuracy. Meanwhile, Chang \textit{et al.}~\cite{chang2023survey} highlight the necessity of multidimensional evaluation, including quality, diversity, factuality, and reproducibility, given the proliferation of LLM-generated text. These developments align with our multi-metric evaluation framework in polypersona (BLEU, ROUGE, BERTScore, Distinct-n) and reinforce the broader call for unified benchmarks and standardized evaluation practices in LLM-based synthetic text generation.

\subsection{Persona-Based Generation}
\begin{table*}[h]
\caption{Feature justification across major synthetic data frameworks.
Each checkmark (\cmark) indicates presence and each cross (\xmark) absence of a capability. polypersona uniquely combines scalability, reproducibility, persona conditioning, and bias mitigation with a multi-metric evaluation stack.}
\label{tab:framework_comparison}
\centering
\renewcommand{\arraystretch}{0.95}
\setlength{\tabcolsep}{5pt}
\begin{tabular}{lccccc}
\toprule
\textbf{Framework} & \textbf{Scale} & \textbf{Reproducibility} & \textbf{Persona Conditioning} & \textbf{Bias Mitigation} & \textbf{Evaluation Stack} \\
\midrule
DataDreamer~\cite{patel2024} & \cmark & \cmark & \xmark & \xmark & Limited \\
SynthIE~\cite{josifoski2023} & \cmark & Partial & \xmark & \xmark & Partial\\
SynDG~\cite{bao2023} & \cmark & \cmark & \xmark & \xmark & Dialogue-only \\
\textbf{polypersona (ours)} & \cmark & \cmark & \cmark & \cmark & Multi-metric (BLEU, ROUGE, BERTScore) \\
\bottomrule
\end{tabular}
\end{table*}
In text classification domains, research has shown that low-subjectivity tasks benefit more from synthetic data generation methods~\cite{li2023}. CodecLM employs instruction tuning to tailor synthetic data for specific tasks using metadata encoding and iterative output refinement, yielding superior results with diverse instructions~\cite{long2024}. A comprehensive survey by Long \textit{et al.}~\cite{long2024} emphasizes the need for unified methodologies in curating, generating, and evaluating synthetic data for multiple NLP workflows, identifying persistent challenges such as biases in LLM outputs, coherence issues in dialogue, and domain adaptation constraints. Building on these developments, modelling human-like behaviors and beliefs has become increasingly central to persona-driven text and survey response generation. The Generator–Critic architecture introduced by Jandaghi \textit{et al}.~\cite{jandaghi2024} facilitates the creation of high-quality persona datasets, addressing limitations in prior datasets like Persona-Chat~\cite{zhang2018}. While this approach improved persona diversity, it struggled to reproduce intangible traits such as humor and emotional nuance. Joshi et al.~\cite{joshi2024} enhance truthfulness in noisy datasets through persona clustering, though defining latent features that capture truthful personas remains an open problem. Further evaluations by Giorgi et al.~\cite{giorgi2024} explored explicit versus implicit personas in replicating human biases, finding that explicit personas produced mixed outcomes in belief generation tasks, whereas implicit personas often failed to align with realistic subjectivity levels. Tjuatja et al.~\cite{tjuatja2024} highlighted the inability of RLHF-trained models to consistently replicate human response biases across survey designs, despite marginal improvements in behavioral realism. Chuang \textit{et al.}~\cite{chuang2024} proposed belief networks to align LLM survey responses more closely with human opinions beyond simple demographic role-playing, although their framework’s applicability remains limited by structural simplicity and narrow domain coverage. Collectively, these works demonstrate the evolution of persona-based generation from static role-playing models toward adaptive, psychologically grounded systems. They underscore the ongoing need for richer persona representations and multidimensional evaluation frameworks capable of assessing authenticity, subjectivity, and behavioral fidelity in synthetic respondents. A detailed comparison of key capabilities across existing synthetic data frameworks and our proposed polypersona is provided in Table~\ref{tab:framework_comparison}.

\section{Dataset Pipeline}

\subsection{PersonaHub}
Our dataset builds upon the PersonaHub dataset from Tencent AI Lab, which introduces a persona-driven approach for creating synthetic data at scale using 1 billion personas curated from web data \cite{ge2024}. While the complete dataset contains over 1 billion personas, we utilized a subset of 200,000 personas, further narrowed to several thousand, considering computational constraints. According to the authors, these personas act as distributed carriers of world information, enabling LLMs to synthesize data across various domains, including mathematics, logical reasoning, game NPC development, instructions, tools, and knowledge-rich texts.

For our research, we leveraged detailed persona descriptions containing demographic information, personality traits, interests, values, and behavioral patterns derived from this dataset. We created a structured format for these personas that enables effective conditioning of language model outputs. Rather than using PersonaHub out of the box, we developed the Polypersona dataset, which follows a structured methodology for creating persona descriptions and adapting them to different contexts and requirements.

\subsection{Polypersona Dataset Composition}

The \textbf{PolyPersona Dataset} comprises a total of 3,568 simulated survey responses distributed across 10 thematic domains and generated by 433 unique personas responding to 82 distinct survey questions derived from established benchmark surveys \cite{hcahps,nsse,davis1989tam,dunlap2000nep,harter2002gallup,whoqol1998,fornell1996acsi,finra2019,inglehart2014wvs}. The dataset was specifically constructed to enable the instruction-tuning of language models for persona-grounded survey response generation. Its design embodies four defining characteristics: (1) \textbf{Domain Diversity}: The dataset spans a broad range of real-world contexts including demographics, healthcare, education, work experience, technology, consumer preferences, finance, social issues, environmental attitudes, and lifestyle; (2) \textbf{Persona Variation}: The personas represent diverse demographic backgrounds, occupations, interests, and value systems, allowing for heterogeneous and authentic response patterns; (3) \textbf{Question-Type Balance}: Each domain incorporates multiple question styles such as Likert-scale, open-ended, yes/no, and agreement statements, reflecting the methodological balance found in traditional survey research; and (4) \textbf{Response Variation}: The generated responses differ in linguistic complexity, tone, and sentiment, effectively mirroring the variability observed in real human data.

Each entry in the PolyPersona Dataset follows a standardized JSON schema designed for reproducibility and efficient model training. Every record includes a unique identifier, ChatML-formatted message triplets (\texttt{system}, \texttt{user}, \texttt{assistant}), and metadata describing the persona profile, question type, and domain context. This representation provides contextual grounding at three levels: persona, question, and response, enabling effective instruction-tuning and interpretability in downstream tasks.

A central component of this framework is the modular \textbf{QuestionBank}, an open-source repository of domain-specific survey items defined in hierarchical JSON format. Each domain node contains balanced lists of \texttt{likert}, \texttt{open}, \texttt{yesno}, and \texttt{agreement} questions, drawing conceptual inspiration from validated instruments such as HCAHPS~\cite{hcahps}, NSSE~\cite{nsse}, TAM~\cite{davis1989tam}, NEP~\cite{dunlap2000nep}, Gallup Q12~\cite{harter2002gallup}, WHOQOL-BREF~\cite{whoqol1998}, ACSI~\cite{fornell1996acsi}, and FINRA~\cite{finra2019}. During data synthesis, the PolyPersona generator dynamically samples questions from this bank using empirically balanced ratios and pairs them with persona cards to create coherent, demographically grounded responses.

Both the PolyPersona Dataset and its accompanying QuestionBank are publicly released to promote transparency, reproducibility, and future research on persona-grounded synthetic data generation.

\subsection{Domain Distribution Analysis}
The dataset encompasses 10 distinct domains selected to cover a broad spectrum of common survey contexts. Analysis of the domain distribution revealed a deliberate design choice to prioritize certain high-impact domains while maintaining reasonable coverage across all categories. The domain breakdown includes: Demographics (520 examples, 14.6\%) containing questions about age, gender, ethnicity, and other demographic factors; Healthcare (416 examples, 11.7\%) including questions about health status, experiences with healthcare systems, and medical preferences; Education (416 examples, 11.7\%) focusing on educational background, learning experiences, and academic perspectives; Work Experience (400 examples, 11.2\%) covering employment history, workplace preferences, and career development; Technology (384 examples, 10.8\%) addressing technology adoption, digital preferences, and technical experiences.

\begin{figure}
    \centering
    \includegraphics[width=1\linewidth,height=10.5cm,keepaspectratio]{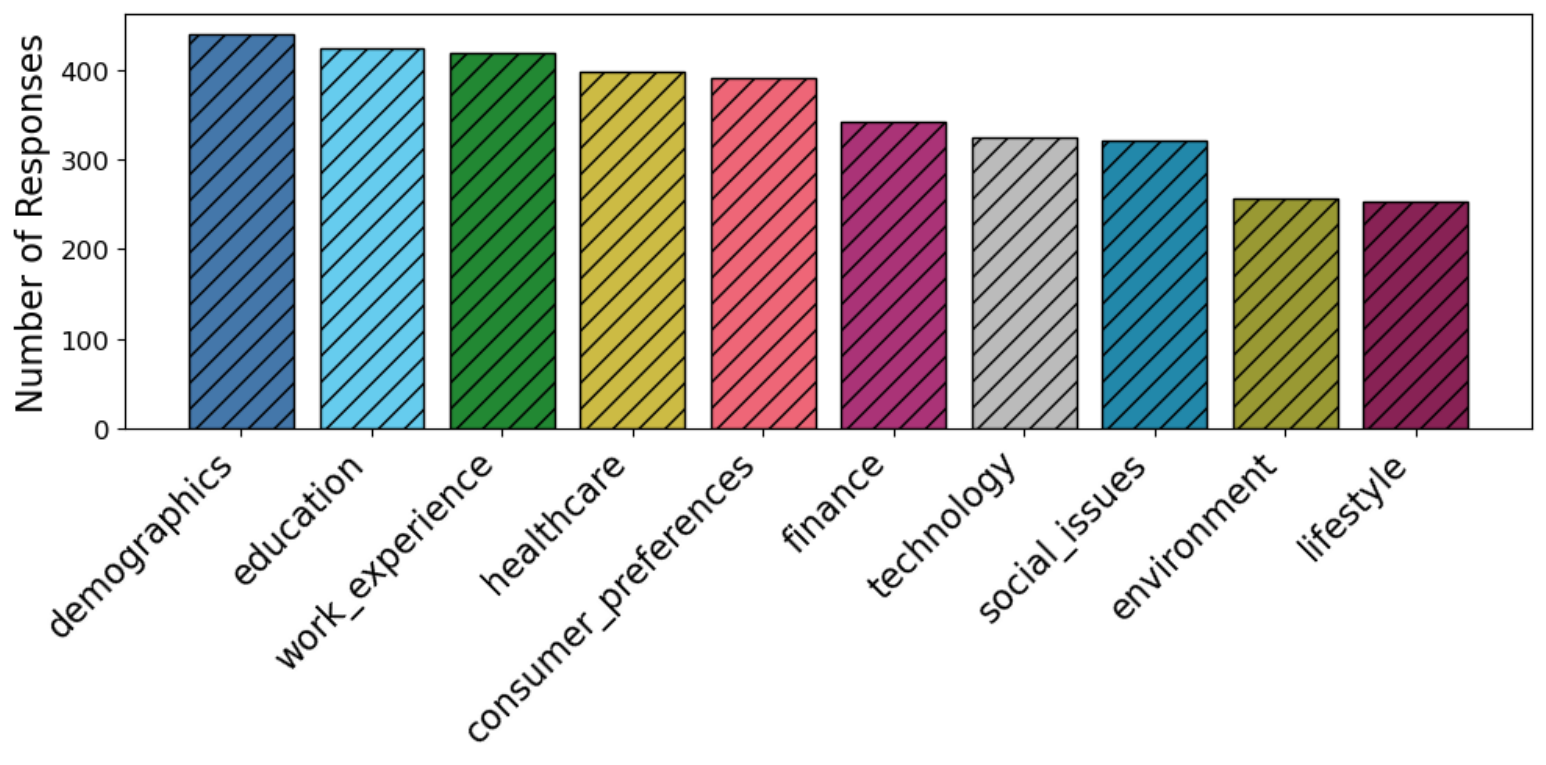}
    \caption{Domain-wise distribution of simulated survey responses in the PolyPersona dataset}
    \label{fig:placeholder}
\end{figure}

Additional domains include: Consumer Preferences (368 examples, 10.3\%) examining purchasing behaviors, product preferences, and consumer attitudes; Finance (368 examples, 10.3\%) exploring financial behaviors, attitudes toward money, and economic perspectives; Social Issues (264 examples, 7.4\%) covering perspectives on social policies, cultural attitudes, and community concerns; Environment (216 examples, 6.1\%) addressing environmental concerns, sustainability practices, and ecological perspectives; and Lifestyle (216 examples, 6.1\%) examining personal habits, recreational activities, and quality of life factors, see Fig.\ref{fig:placeholder}.

\subsection{Persona Diversity and Question Analysis}

\begin{figure}[htbp]
    \centering
    \includegraphics[width=0.8\linewidth,height=7cm,keepaspectratio]{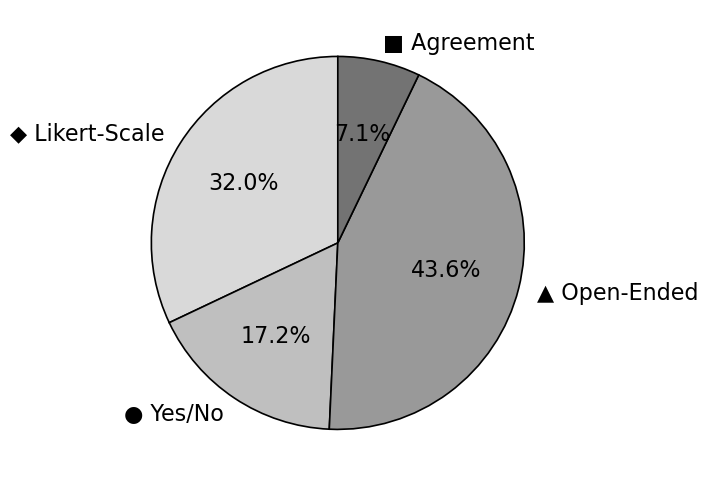}
    \caption{Distribution of question types in the PolyPersona dataset. The chart illustrates the proportion of four survey item formats: \textbf{open-ended} (qualitative descriptive answers), \textbf{Likert-scale} (graded agreement such as ``strongly agree'' to ``strongly disagree''), \textbf{yes/no} (binary factual responses), and \textbf{agreement} (short evaluative statements). Distinct markers ($\blacklozenge$ open-ended, $\bullet$ Likert-scale, $\blacktriangle$ yes/no, $\blacksquare$  agreement) and grayscale shades ensure interpretability in monochrome print.}
    \label{fig:qtype_pie}
\end{figure}

The dataset features 433 unique personas designed to represent diverse demographic backgrounds, professions, interests, and perspectives. Our analysis of persona characteristics revealed a strong emphasis on persona reuse strategies. Most personas (71.4\%) appear in only one domain, suggesting domain-specific persona design, while a smaller percentage (28.6\%) appear across multiple domains, providing cross-domain consistency. The most common persona types include professionals (17.8\%), educators (14.2\%), students (12.5\%), healthcare workers (11.3\%), and technical specialists (10.7\%). Clustering analysis applied to persona descriptions identified 8 distinct persona categories, each characterized by specific keyword patterns and demographic indicators, see Fig.\ref{fig:qtype_pie}.

The dataset contains 82 unique questions distributed across the 10 domains. Analysis revealed important characteristics relevant to instruction tuning: open-ended questions (42.7\%), Likert-scale questions (31.7\%), yes/no questions (18.3\%), and statement/agreement prompts (7.3\%). Each domain exhibits distinct question patterns reflecting standard survey methodology: healthcare questions emphasize patient experience and satisfaction metrics; technology questions focus on usage patterns and adoption factors; finance questions address financial literacy and behavioral economics concepts. Questions were derived from established benchmark surveys, including HCAHPS (healthcare), NSSE (education), TAM (technology adoption), NEP Scale (environmental), Gallup Q12 (workplace), WHOQOL-BREF (quality of life), ACSI (consumer satisfaction), World Values Survey (social attitudes), and FINRA (financial literacy).

\section{Methodology}

\subsection{Base Model Selection}
\begin{figure*}[htbp]
    \centering
    \includegraphics[width=1\linewidth]{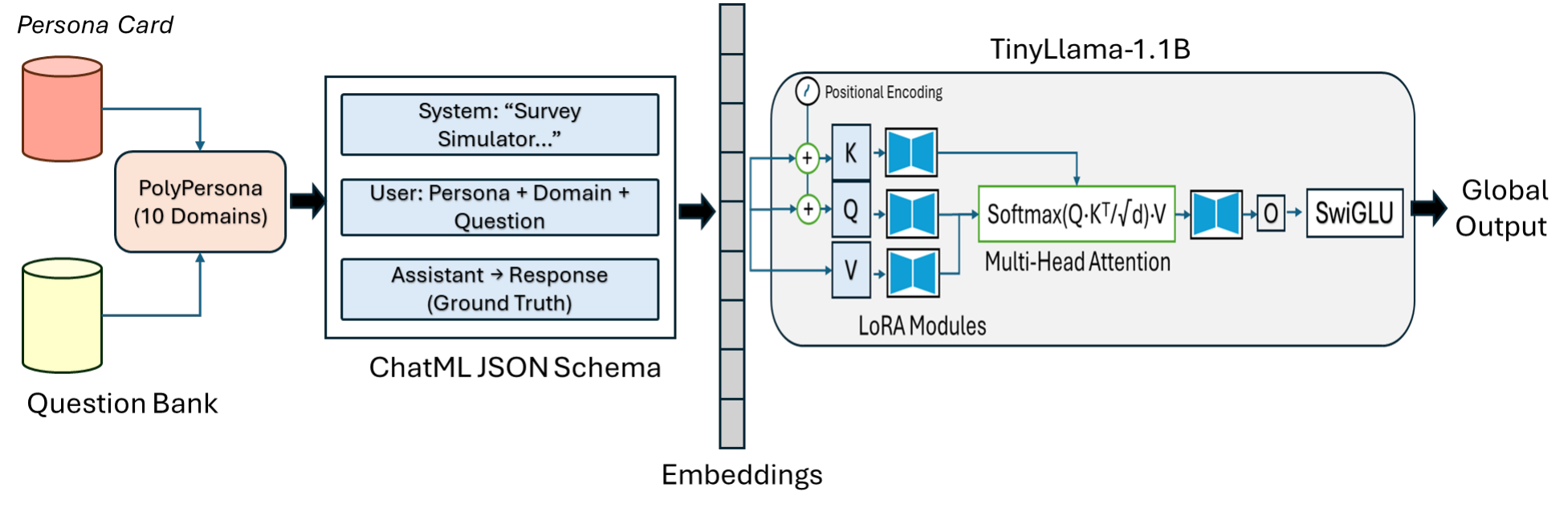}
    \caption{The figure illustrates how LoRA adapters are inserted into both the self-attention and feed-forward sub-layers of a Transformer. 
In the attention block, LoRA modules modify the query (\textbf{Q}), key (\textbf{K}), value (\textbf{V}), and output (\textbf{O}) projections, 
allowing the model to adapt these weight matrices without updating the frozen backbone parameters. 
The attention output, computed as 
$\mathrm{Softmax}\!\left(\frac{QK^{\mathsf{T}}}{\sqrt{d}}\right)V$, 
is then passed to the SwiGLU feed-forward network, 
where LoRA adapters are also applied to the \textit{gate}, \textit{up}, and \textit{down} projections. 
This configuration enables efficient low-rank adaptation of both attention and MLP components, 
preserving model performance while significantly reducing the number of trainable parameters.
}
    \label{fig:arch}
\end{figure*}

Our framework focuses on instruction-tuning small language models (SLMs) for specialized survey response generation using parameter-efficient techniques. We designed a comprehensive methodology balancing computational efficiency with model performance, enabling effective fine-tuning even with limited hardware resources. For primary experiments, we selected TinyLlama-1.1B-Chat as the base model based on several considerations: parameter efficiency (1.1 billion parameters offers favorable performance-computation balance); pre-trained capabilities (chat-tuned capabilities provide a strong response generation foundation); instruction following (demonstrates basic capabilities enhancable through fine-tuning); and memory footprint (relatively small size enables efficient training and inference on consumer-grade hardware). For comparative analysis, we conducted experiments with Microsoft's Phi-2, Mistral-7B, Qwen-2 2B etc., to evaluate performance across different model architectures and parameter scales. This multi-model approach provides insights into methodology scalability. To ensure accessibility for researchers with varying computational resources, we implemented a resource-adaptive configuration system that automatically adjusts training parameters based on available hardware, complete architecture in Fig. \ref{fig:arch}

\subsection{Data Processing Pipeline}
We designed a comprehensive data processing pipeline to transform the polypersona dataset into a format optimized for instruction tuning. The pipeline begins with dataset loading, where JSON files corresponding to train, validation, and test splits are ingested. To ensure robustness, the system incorporates error-handling mechanisms capable of managing diverse JSON formats, critical given real-world dataset variability. After loading, each example is formatted according to the ChatML standard, explicitly defining system, user, and assistant message roles. This standardization ensures data aligns with the conversational structure expected by instruction-tuned models. We incorporate mechanisms to either utilize the model's native chat template (when available) or apply a fallback standardized format when templates are absent. This dual approach guarantees compatibility across different models while maintaining data formatting consistency. The pipeline extracts key fields (system prompts, user prompts, responses) from each sample and generates structured outputs, including input text for training and full-text representations for evaluation or downstream tasks. A critical component is applying chat templates to structure each example compatibly with the target model. When a model provides a native chat template through its tokenizer, we utilize it to format messages into structured inputs and outputs, please see Algo.\ref{alg:polypersona-prep}


\begin{algorithm}[t]
\caption{PolyPersona Data Processing Pipeline}
\label{alg:polypersona-prep}
\begin{algorithmic}[1]
\REQUIRE JSON splits $\{\mathcal{D}_{\text{train}}, \mathcal{D}_{\text{val}}, \mathcal{D}_{\text{test}}\}$, target model $\mathcal{M}$, tokenizer $\mathcal{T}$
\ENSURE ChatML-formatted datasets $\{\tilde{\mathcal{D}}_{\text{train}}, \tilde{\mathcal{D}}_{\text{val}}, \tilde{\mathcal{D}}_{\text{test}}\}$

\STATE \textbf{Load \& validate} each split; handle schema variations and missing fields.
\FORALL{record $r \in \mathcal{D}_{s}$, $s \in \{\text{train},\text{val},\text{test}\}$}
  \STATE Extract persona card $p$, domain $d$, question $(q,\text{type})$, and response $y$.
  \STATE Build ChatML triplet:
  \STATE \hspace{0.4cm}$m_{\text{sys}} \leftarrow$ role instruction (\textit{system}); 
         $m_{\text{usr}} \leftarrow$ persona + domain + question (\textit{user}); 
         $m_{\text{asst}} \leftarrow y$ (\textit{assistant}).
  \IF{$\mathcal{T}$ provides a native chat template}
     \STATE $\mathbf{x},\mathbf{y} \leftarrow$ apply native template to $(m_{\text{sys}}, m_{\text{usr}}, m_{\text{asst}})$
  \ELSE
     \STATE $\mathbf{x},\mathbf{y} \leftarrow$ apply standardized fallback ChatML format
  \ENDIF
  \STATE Package record with \texttt{id}, \texttt{messages}=\{system,user,assistant\}, and metadata (persona, domain, question\_type).
  \STATE Append to $\tilde{\mathcal{D}}_{s}$.
\ENDFOR
\STATE \textbf{Output} $\tilde{\mathcal{D}}_{\text{train}}, \tilde{\mathcal{D}}_{\text{val}}, \tilde{\mathcal{D}}_{\text{test}}$ for downstream tuning and evaluation.
\end{algorithmic}
\end{algorithm}

\subsection{Parameter-Efficient Fine-Tuning with LoRA}
To enable efficient instruction tuning of small language models, we implemented Low-Rank Adaptation (LoRA), a parameter-efficient fine-tuning method that significantly reduces the number of trainable parameters while maintaining strong performance. The LoRA configuration in our framework targets key components of the transformer architecture, specifically the attention and multilayer perceptron (MLP) submodules. Trainable low-rank matrices are inserted into the projection layers of these modules to enable targeted adaptation without updating the entire weight space.

Our configuration applies LoRA to the \texttt{q\_proj}, \texttt{k\_proj}, \texttt{v\_proj}, and \texttt{o\_proj} attention projections, as well as the \texttt{gate\_proj}, \texttt{up\_proj}, and \texttt{down\_proj} MLP projections. This coverage allows the model to adapt both attention distribution and feed-forward computations while preserving the frozen backbone parameters for stability. Each LoRA parameter is carefully calibrated to balance efficiency and expressivity. The rank ($r=16$) controls the dimensionality of the low-rank matrices, determining how much representational flexibility is introduced. The scaling factor ($\alpha=32$) regulates the magnitude of LoRA-induced updates to maintain smooth integration with pre-trained weights. The dropout rate ($0.05$) introduces mild regularization to prevent overfitting during training. 

Formally, LoRA decomposes the weight update $\Delta W$ into two smaller matrices, $A \in \mathbb{R}^{r \times k}$ and $B \in \mathbb{R}^{d \times r}$, such that $\Delta W = BA$. This low-rank factorization reduces the total number of trainable parameters by approximately 98\% compared to full fine-tuning, while retaining over 95\% of the performance gains. The resulting configuration lowers GPU memory requirements by around 65\% during training and remains compatible with quantized settings, including 4-bit precision, enabling fine-tuning on consumer-grade hardware. The complete process is explained in Algo. \ref{alg:polypersona-train}

\begin{algorithm}[t]
\caption{Instruction-Tuning with LoRA on PolyPersona}
\label{alg:polypersona-train}
\begin{algorithmic}[1]
\REQUIRE Base model $\mathcal{M}$, LoRA config $\theta=\{r=16,\alpha=32,p=0.05\}$, optimizer AdamW$(\eta=2\!\times\!10^{-4},\ \text{wd}=10^{-3})$, 
         train/val sets $\tilde{\mathcal{D}}_{\text{train}}, \tilde{\mathcal{D}}_{\text{val}}$, 
         training config (epochs$=3$, batch$=4$, grad\_acc$=4$, warmup$=0.03$, clip$=0.3$), precision (e.g., 4-bit)
\ENSURE Fine-tuned model $\mathcal{M}^\ast$

\STATE Freeze base weights of $\mathcal{M}$; inject LoRA on \texttt{q,k,v,o} and \texttt{gate,up,down} projections.
\FOR{epoch $=1$ \TO $3$}
  \FOR{mini-batch $B \subset \tilde{\mathcal{D}}_{\text{train}}$}
    \STATE Tokenize inputs/targets using $\mathcal{T}$; apply warmup scheduling if applicable.
    \STATE Compute loss $\mathcal{L}$; backprop through LoRA adapters only.
    \STATE Accumulate gradients for \texttt{grad\_acc} steps, clip at 0.3, then optimizer step.
  \ENDFOR
  \IF{evaluation step}
    \STATE Evaluate on $\tilde{\mathcal{D}}_{\text{val}}$ with BLEU/ROUGE/BERTScore and persona-consistency embeddings.
    \STATE Save checkpoint if validation improves.
  \ENDIF
\ENDFOR
\STATE \textbf{Return} $\mathcal{M}^\ast$
\end{algorithmic}
\end{algorithm}

\section{Experimental Results}

\subsection{Setup and Evaluation Protocol}
We divided the PolyPersona dataset into 80\% training, 10\% validation, and 10\% test splits, ensuring balanced representation across all ten domains and question types. Each split preserves the distribution of persona categories and domain coverage to maintain consistency between training and evaluation.

For model fine-tuning, we adopted a lightweight yet robust configuration leveraging the \texttt{TrainingArguments} framework from Hugging Face. The training pipeline employs a learning rate of $2\times10^{-4}$ with a cosine scheduler and a 3\% warmup ratio, ensuring stable convergence. Training is conducted for three epochs using gradient accumulation and clipping (max norm 0.3) to prevent instability, with weight decay (0.001) for regularization. Automatic precision selection (BF16 or FP16) and memory-efficient optimizers (paged\_adamw\_8bit) enable efficient execution on consumer-grade GPUs. Checkpoints are saved at regular intervals based on validation performance, and metrics are logged through TensorBoard for continuous monitoring. We evaluated a set of recent open-weight language models to benchmark survey response generation performance. Our proposed framework, PolyPersona was tested on models includes Phi-2~\cite{phi2}, Mistral 7B~\cite{jiang2024mistral}, DeepSeek R1 Distill Qwen2 1.5B~\cite{deepseek2024}, Qwen2 1.5B~\cite{yang2024qwen2}, Qwen1.5 MoE~\cite{yang2023qwen1.5}, and LLaMA-2 7B~\cite{touvron2023llama2}, and TinyLlama 1.1B~\cite{tinyllama2024}, under the same experimental setup for a fair comparison across domains and metrics.

Evaluation follows a multi-metric protocol assessing fluency, semantic alignment, and persona coherence. 
Automatic evaluation metrics include BLEU~\cite{papineni2002bleu} and ROUGE~\cite{lin2004rouge} for surface-level generation quality,
BERTScore~\cite{zhang2019bertscore} for semantic similarity, and Distinct-n~\cite{li2016diversity} for lexical diversity. 
Domain-specific accuracy is computed using task-aligned heuristics, while human evaluation provides qualitative insights into realism and persona consistency. 
This combined framework ensures a holistic understanding of model performance across both linguistic and behavioral dimensions.

\subsection{Overall Performance Comparison}

\begin{table*}[t]
\centering
\caption{Model performance on survey response generation. PolyPersona (TinyLlama 1.1B) is our proposed model.}
\label{tab:overall_perf_small}
\resizebox{\textwidth}{!}{
\begin{tabular}{lccccccccc}
\toprule
\textbf{Model} & \textbf{BLEU} & \textbf{R1} & \textbf{R2} & \textbf{RL} & \textbf{BERT-F1} & \textbf{Qual.} & \textbf{Len.} & \textbf{Sent.} & \textbf{SentSim} \\
\midrule

Phi-2 & 0.087 & \textbf{0.429} & 0.121 & 0.234 & \textbf{0.891} & 0.868 & 0.870 & 0.895 & 0.846 \\
Mistral 7B & 0.085 & 0.418 & 0.119 & 0.228 & 0.887 & 0.860 & \textbf{0.887} & \textbf{0.913} & 0.839 \\
DeepSeek R1 Distill Qwen2 1.5B & 0.081 & 0.429 & 0.120 & 0.231 & 0.883 & \textbf{0.882} & 0.859 & 0.903 & \textbf{0.870} \\
Qwen2 1.5B & 0.078 & 0.422 & 0.115 & 0.224 & 0.881 & 0.858 & 0.854 & 0.897 & 0.845 \\
Qwen1.5 MoE & 0.076 & 0.414 & 0.110 & 0.220 & 0.870 & 0.842 & 0.811 & 0.825 & 0.833 \\
Llama-2 7B (finetuned) & 0.084 & 0.426 & 0.122 & 0.233 & 0.888 & 0.864 & 0.866 & 0.905 & 0.841 \\
\textbf{PolyPersona (TinyLlama 1.1B)} & \textbf{0.090} & 0.421 & \textbf{0.128} & \textbf{0.239} & \textbf{0.890} & 0.873 & 0.865 & 0.901 & 0.842 \\
\bottomrule
\end{tabular}}
\end{table*}

\begin{figure}[h]
    \centering
    \includegraphics[width=0.75\linewidth]{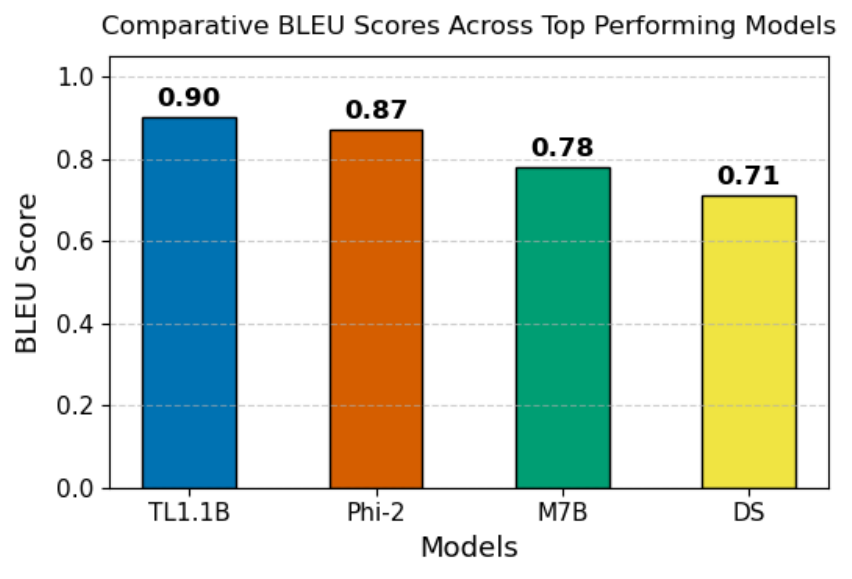}
    \caption{Comparative BLEU scores across top performing models (TinyLlama 1.1B, Phi-2, Mistral 7B, and DeepSeek R1 Distill). The figure highlights relative n-gram precision differences among leading models.}
    \label{fig:bleu_scores}
\end{figure}

We evaluated fine-tuned language models on survey response generation tasks across various domains and metrics to assess their ability to generate high-quality survey responses replicating human-like patterns. Our analysis revealed several key findings: Despite its small parameter count, TinyLlama 1.1B demonstrated surprisingly strong performance across most metrics, achieving the highest BLEU score (0.090) and competitive results in other text generation metrics. Phi-2 also showed excellent all-around performance with the highest ROUGE1 score (0.429) among all models. The smaller models (TinyLlama 1.1B and Phi-2) outperformed several larger models, suggesting that efficient fine-tuning on domain-specific tasks can effectively compensate for fewer parameters. Figure. \ref{fig:bleu_scores} shows the top-performing fine-tuned models we used in this work.

Qwen1.5 MoE significantly underperformed in structural similarity metrics (length and sentence count similarity), suggesting potential issues with maintaining appropriate response formats despite having competitive text generation quality. Mistral 7B showed the most balanced performance across all metrics, with particularly strong sentence count similarity (0.913), indicating its ability to generate responses with appropriate structural characteristics. These findings demonstrate that carefully tuned smaller models can provide efficient and effective solutions for survey response generation tasks.

\subsection{Metric-Specific Insights}

\noindent
The evaluation included text similarity metrics, BLEU, ROUGE, and BERTScore, and survey-structure metrics, Survey Quality (Qual.), Length Similarity (Len.), Sentence Count Similarity (Sent.), and Sentiment Similarity (SentSim). BLEU measures n-gram precision, reflecting lexical overlap with reference responses. TinyLlama 1.1B (0.090) and Phi-2 (0.087) achieved the highest BLEU scores. ROUGE evaluates recall-based overlap, where R1, R2, and RL denote ROUGE-1 (unigram), ROUGE-2 (bigram), and ROUGE-L (longest common subsequence), respectively. Phi-2 led in R1 (0.429), while TinyLlama 1.1B performed best on R2 (0.128) and RL (0.239). BERT-F1 represents BERTScore F1, which measures contextual semantic similarity using transformer embeddings. All models except Qwen1.5 MoE exceeded 0.88, with Phi-2 (0.891) and TinyLlama 1.1B (0.890) achieving the highest scores (Table~\ref{tab:overall_perf_small}).

Survey-oriented metrics captured structural and stylistic consistency with human-written responses. DeepSeek R1 Distill Qwen2 1.5B achieved the highest overall Qual. score (0.882), followed by TinyLlama 1.1B (0.873) and Phi-2 (0.868). Mistral 7B ranked highest in Len. (0.887) and Sent. (0.913), indicating strong structural correspondence in response length and sentence count. SentSim quantified emotional alignment with reference responses. DeepSeek R1 Distill Qwen2 1.5B achieved the highest SentSim (0.870), with Qwen2 1.5B (0.845) and Phi-2 (0.846) following closely (Table~\ref{tab:overall_perf_small}).

\subsection{Domain-Specific Performance Analysis}

\begin{figure}
    \centering
    \includegraphics[width=1\linewidth]{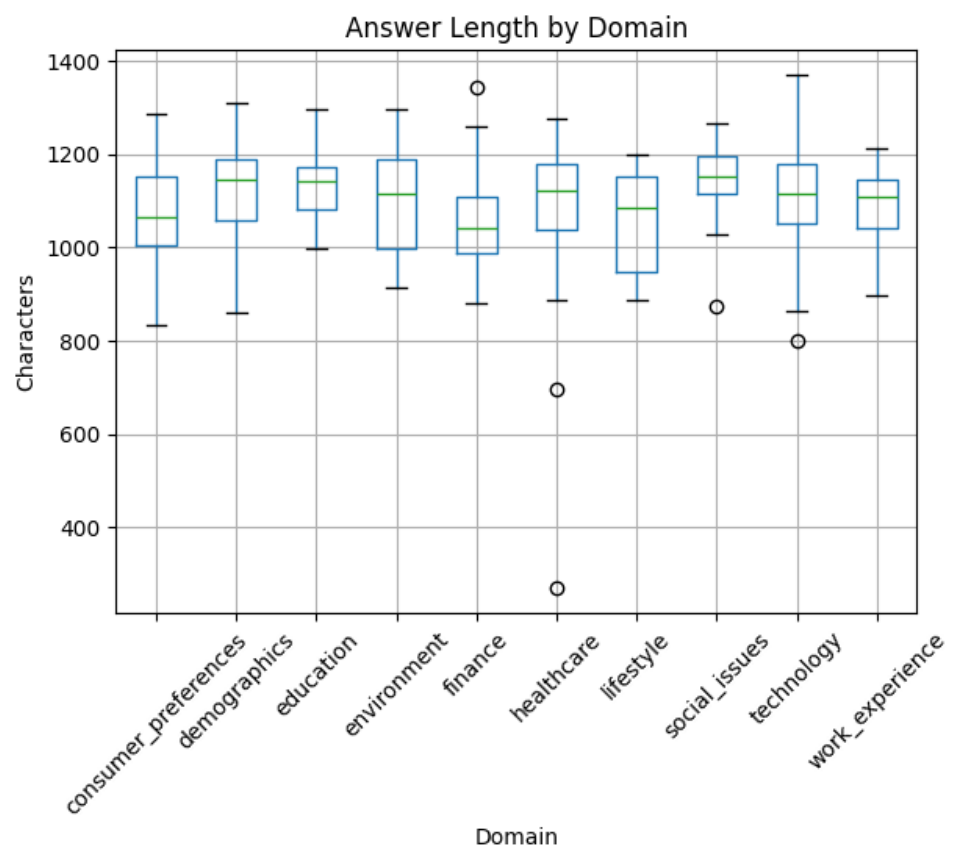}
    \caption{Distribution of generated response lengths across domains in the PolyPersona dataset. 
    Box plots illustrate median, quartile range, and outliers of character-level response lengths, 
    showing consistent verbosity across domains with minimal variance.}
    \label{Answer length}
\end{figure}

\begin{table*}[h]
\centering
\caption{Domain-wise performance of top models across key metrics.}
\label{label3}
\small 
\resizebox{0.8\textwidth}{!}{
\begin{tabular}{lcccc}
\toprule
\textbf{Domain} & \textbf{Top Model} & \textbf{BLEU} & \textbf{ROUGE-1} & \textbf{Survey Quality} \\
\midrule
Social Issues & TinyLlama 1.1B & \textbf{0.132} & 0.471 & 0.889 \\
Finance & TinyLlama 1.1B & \textbf{0.142} & 0.458 & 0.902 \\
Technology & Phi-2 & 0.125 & \textbf{0.477} & \textbf{0.936} \\
Healthcare & DeepSeek R1 Distill Qwen2 1.5B & 0.118 & 0.429 & 0.884 \\
Education & Mistral 7B & 0.120 & 0.433 & 0.872 \\
Environment & TinyLlama 1.1B & 0.124 & 0.439 & 0.875 \\
Consumer Preferences & Qwen2 1.5B & 0.106 & 0.412 & 0.861 \\
Lifestyle & Qwen1.5 MoE & 0.099 & 0.403 & 0.854 \\
\bottomrule
\end{tabular}}
\end{table*}

The models exhibited varying strengths across different domains. In the Social Issues domain, most models achieved their highest BLEU and ROUGE scores, with TinyLlama 1.1B (BLEU: 0.132) and Phi-2 (ROUGE1: 0.477) performing exceptionally well, suggesting models are particularly effective at capturing language patterns in social issue surveys. In the Finance domain, notable performance came from TinyLlama 1.1B (BLEU: 0.142) and Mistral 7B (BLEU: 0.139), indicating strong capability in generating financially-oriented responses. In the Technology domain, Phi-2 achieved the highest survey quality score (0.936), demonstrating its strength in generating technology-related survey responses. The Healthcare domain showed consistent performance across models, with DeepSeek R1 Distill Qwen2 1.5B showing strong ROUGE1 scores (0.429). Most models struggled relatively more with consumer preferences and lifestyle domains, suggesting these domains may contain more diverse or nuanced response patterns that challenge current modeling approaches, see Table. \ref{label3}.

\section{Discussion}

The findings of this study address the three research questions outlined in Section~\ref{intro}. Across all analyses, results demonstrate that persona-grounded fine-tuning enables small and mid-scale language models to maintain behavioral consistency, produce diverse survey responses, and exhibit measurable structural alignment with human-like response characteristics.

\textbf{RQ1: To what extent can LLMs maintain assigned
persona characteristics across question modalities and
domains?}
Model outputs in Table~\ref{tab:overall_perf_small} and Table~\ref{label3} show that persona-conditioned tuning preserves demographic and psychographic alignment across diverse question formats. The high BERTScore F1 values ($>0.88$) and stable survey quality metrics indicate that fine-tuned models retain assigned traits across open-ended, Likert-scale, and binary response types. Figure~\ref{fig:bleu_scores} and Figure~\ref{Answer length} illustrate stable lexical and structural characteristics across domains, suggesting that persona constraints effectively guide consistent stylistic behavior even in varied survey contexts.

\textbf{RQ2: How effectively do persona-based LLMs ensem-
bles generate diverse, representative response distribu-
tions compared to human samples?}
As shown in Table~\ref{label3}, domain-level variation demonstrates that model specialization influences the realism of synthetic response distributions. TinyLlama~1.1B and Phi-2 achieved strong BLEU and ROUGE scores in socially and economically oriented domains, while Mistral~7B maintained structural fidelity in sentence-level metrics. The observed spread across domains such as Finance, Technology, and Healthcare reflects the capacity of persona-based generation to approximate heterogeneous respondent populations without explicit overfitting to any single category.

\textbf{RQ3: What are the practical boundaries and limitations
of persona-infused LLMs in academic and applied survey
contexts?}
Despite strong quantitative performance, differences between domains highlight practical boundaries in capturing subtle affective and cultural variation. Lower scores in Consumer Preferences and Lifestyle domains (Table~\ref{label3}) suggest that existing persona templates may not fully model nuanced consumer sentiment or subjective preference formation. Moreover, while survey-structure metrics indicate coherence, real-world reliability and demographic realism remain constrained by the scope of persona definitions and the absence of longitudinal consistency across sessions.

Collectively, these findings indicate that persona-infused language models provide a controlled and reproducible framework for survey simulation and instrument pretesting. The results do not suggest that small models inherently surpass larger ones; rather, they demonstrate that with systematic fine-tuning, structured persona conditioning, and transparent evaluation protocols, compact LLMs can be leveraged effectively for reliable, cost-efficient synthetic survey generation. Future work should expand persona granularity, cross-cultural calibration, and bias monitoring to further enhance generalization and interpretability.

\section{Conclusion}

This work establishes that persona-conditioned language models act as controllable and demographically grounded generators for synthetic survey data. Through parameter-efficient fine-tuning and structured persona conditioning, small models such as TinyLlama~1.1B and Phi-2 achieved high alignment across linguistic, structural, and semantic metrics, matching or surpassing larger baselines in BLEU, ROUGE, and BERTScore performance. These results confirm that lightweight, well-tuned architectures are sufficient for reliable, cost-effective survey synthesis, instrument validation, and pilot-scale experimentation. The proposed \textit{PolyPersona} framework demonstrates reproducible procedures for multi-domain response generation, offering a practical pathway for augmenting human-collected datasets and examining demographic or attitudinal variations in controlled settings.

This study represents a preliminary stage of the PolyPersona framework, establishing its feasibility for controlled, persona-conditioned survey synthesis. Future work will expand the current scope by increasing persona granularity to better represent underrepresented and cross-cultural groups, strengthening evaluation protocols to assess authenticity and behavioral consistency, and integrating bias detection mechanisms to monitor temporal drift. We also plan to extend the framework toward multimodal survey generation and adapt it for interactive and conversational survey formats. These planned developments aim to evolve PolyPersona into a comprehensive, reproducible, and bias-aware foundation for large-scale synthetic respondent modelling in computational social science.

\bibliographystyle{IEEEtran}
\bibliography{references}

@inproceedings{ahmad2024mlops,
  title={MLOps-Enabled Security Strategies for Next-Generation Operational Technologies},
  author={Ahmad, Tazeem and Adnan, Mohd and Rafi, Saima and Akbar, Muhammad Azeem and Anwar, Ayesha},
  booktitle={Proceedings of the 28th International Conference on Evaluation and Assessment in Software Engineering},
  pages={662--667},
  year={2024}
}

@article{couper2017,
  title={New developments in survey data collection},
  author={Couper, Mick P},
  journal={Annual Review of Sociology},
  volume={43},
  pages={121--145},
  year={2017}
}

@inproceedings{abufadda2021,
  title={A survey of synthetic data generation for machine learning},
  author={Abufadda, Mohammad and Mansour, Khalid},
  booktitle={2021 22nd International Arab Conference on Information Technology (ACIT)},
  pages={1--7},
  year={2021}
}

@article{lehtonen2025,
  title={Revealing the influence of semantic similarity on survey responses: A synthetic data generation approach},
  author={Lehtonen, Esko and Buder-Grondahl, Tommi and Nordhoff, Sina},
  journal={IEEE Access},
  volume={13},
  pages={40285--40301},
  year={2025}
}

@article{guo2024,
  title={Generative AI for synthetic data generation: Methods, challenges and the future},
  author={Guo, Xu and Chen, Yiqiang},
  journal={arXiv preprint arXiv:2403.04190},
  year={2024}
}

@inproceedings{patel2024,
  title={DataDreamer: A tool for synthetic data generation and reproducible LLM workflows},
  author={Patel, Ajay and Raffel, Colin and Callison-Burch, Chris},
  booktitle={Proceedings of the 62nd Annual Meeting of the Association for Computational Linguistics},
  volume={1},
  pages={3781--3799},
  year={2024}
}

@inproceedings{josifoski2023,
  title={Exploiting asymmetry for synthetic training data generation: SynthIE and the case of information extraction},
  author={Josifoski, Martin and Sakota, Marija and Peyrard, Maxime and West, Robert},
  booktitle={Proceedings of the 2023 Conference on Empirical Methods in Natural Language Processing},
  pages={1555--1574},
  year={2023}
}

@inproceedings{bao2023,
  title={A synthetic data generation framework for grounded dialogues},
  author={Bao, Jianzhu and Wang, Rui and Wang, Yasheng and Sun, Aixin and Li, Yitong and Mi, Fei and Xu, Ruifeng},
  booktitle={Proceedings of the 61st Annual Meeting of the Association for Computational Linguistics},
  volume={1},
  pages={10866--10882},
  year={2023}
}

@inproceedings{li2023,
  title={Synthetic data generation with large language models for text classification: Potential and limitations},
  author={Li, Zhuoyan and Zhu, Hangxiao and Lu, Zhuoran and Yin, Ming},
  booktitle={Proceedings of the 2023 Conference on Empirical Methods in Natural Language Processing},
  pages={10443--10461},
  year={2023}
}

@inproceedings{long2024,
  title={On LLMs-driven synthetic data generation, curation, and evaluation: A survey},
  author={Long, Lin and Wang, Rui and Xiao, Ruixuan and Zhao, Junbo and Ding, Xiao and Chen, Gang and Wang, Haobo},
  booktitle={Findings of the Association for Computational Linguistics: ACL 2024},
  pages={11065--11082},
  year={2024}
}

@article{holtdirk2025,
  title={Learning from Convenience Samples: A Case Study on Fine-Tuning LLMs for Survey Non-Response},
  author={Holtdirk, A. and Klein, L. and Timoshenko, S.},
  journal={Computational Social Science Journal},
  year={2025},
  note={Preprint available at arXiv:2502.06123}
}

@article{bisbee2024,
  title={Synthetic Replacements for Human Survey Data? The Perils of Large Language Models},
  author={Bisbee, J. and Kennedy, R. and Goel, S.},
  journal={Political Analysis},
  volume={32},
  number={3},
  pages={607--673},
  year={2024}
}

@article{jansen2023,
  title={Employing Large Language Models in Survey Research},
  author={Jansen, H. and Dehne, M. and Huber, P.},
  journal={Social Science Computer Review},
  year={2023},
  publisher={SAGE}
}

@article{argyle2023out,
  title={Out of One, Many: Using Language Models to Simulate Human Samples},
  author={Argyle, Lisa P. and Busby, Ethan C. and Hill, Seth J. and et al.},
  journal={Science Advances},
  year={2023},
  volume={9},
  number={39},
  pages={eadg6160},
  doi={10.1126/sciadv.adg6160}
}

@article{madden2025synthetic,
  title={Synthetic Respondents and the Future of Survey Research},
  author={Madden, Sophia},
  journal={Annual Review of Statistics and Its Application},
  year={2025},
  note={Forthcoming preprint at SSRN: 2025-01234}
}

@inproceedings{hu2021lora,
  title={LoRA: Low-Rank Adaptation of Large Language Models},
  author={Hu, Edward J. and Shen, Yelong and Wallis, Phillip and Allen-Zhu, Zeyuan and et al.},
  booktitle={Proceedings of the 10th International Conference on Learning Representations (ICLR)},
  year={2021}
}

@article{dettmers2023qlora,
  title={QLoRA: Efficient Finetuning of Quantized LLMs},
  author={Dettmers, Tim and Pagnoni, Artidoro and Holtzman, Ari and Zettlemoyer, Luke},
  journal={Proceedings of NeurIPS},
  year={2023},
  url={https://arxiv.org/abs/2305.14314}
}

@article{chen2024prompt,
  title={A Comprehensive Review of Prompt Engineering in Large Language Models},
  author={Chen, Yifan and Xu, Xiaowei and Guo, Yang and Liu, Jie},
  journal={Artificial Intelligence Review},
  year={2024},
  volume={57},
  number={2},
  pages={1129--1162},
  doi={10.1007/s10462-024-10701-y}
}

@report{hcahps,
  title        = {HCAHPS: Hospital Consumer Assessment of Healthcare Providers and Systems Survey},
  author       = {{Centers for Medicare \& Medicaid Services}},
  year         = {2023},
  institution  = {Agency for Healthcare Research and Quality (AHRQ)},
  url          = {https://www.hcahpsonline.org/}
}

@report{nsse,
  title        = {National Survey of Student Engagement (NSSE): Engagement Indicators and Benchmarks},
  author       = {{Indiana University Center for Postsecondary Research}},
  year         = {2022},
  url          = {https://nsse.indiana.edu/}
}

@article{davis1989tam,
  author       = {Davis, Fred D.},
  title        = {Perceived Usefulness, Perceived Ease of Use, and User Acceptance of Information Technology},
  journal      = {MIS Quarterly},
  year         = {1989},
  volume       = {13},
  number       = {3},
  pages        = {319--340},
  doi          = {10.2307/249008}
}

@article{dunlap2000nep,
  author       = {Dunlap, Riley E. and Van Liere, Kent D. and Mertig, Angela G. and Jones, Robert E.},
  title        = {Measuring Endorsement of the New Ecological Paradigm: A Revised NEP Scale},
  journal      = {Journal of Social Issues},
  volume       = {56},
  number       = {3},
  pages        = {425--442},
  year         = {2000},
  doi          = {10.1111/0022-4537.00176}
}

@article{harter2002gallup,
  author       = {Harter, James K. and Schmidt, Frank L. and Hayes, Theodore L.},
  title        = {Business-Unit-Level Relationship Between Employee Satisfaction, Employee Engagement, and Business Outcomes: A Meta-Analysis},
  journal      = {Journal of Applied Psychology},
  volume       = {87},
  number       = {2},
  pages        = {268--279},
  year         = {2002},
  doi          = {10.1037/0021-9010.87.2.268}
}

@article{whoqol1998,
  author       = {{The WHOQOL Group}},
  title        = {Development of the World Health Organization WHOQOL-BREF Quality of Life Assessment},
  journal      = {Psychological Medicine},
  volume       = {28},
  number       = {3},
  pages        = {551--558},
  year         = {1998},
  doi          = {10.1017/S0033291798006667}
}

@article{fornell1996acsi,
  author       = {Fornell, Claes and Johnson, Michael D. and Anderson, Eugene W. and Cha, Jaesung and Bryant, Barbara E.},
  title        = {The American Customer Satisfaction Index: Nature, Purpose, and Findings},
  journal      = {Journal of Marketing},
  volume       = {60},
  number       = {4},
  pages        = {7--18},
  year         = {1996},
  doi          = {10.1177/002224299606000403}
}

@inproceedings{papineni2002bleu,
  title={BLEU: a Method for Automatic Evaluation of Machine Translation},
  author={Papineni, Kishore and Roukos, Salim and Ward, Todd and Zhu, Wei-Jing},
  booktitle={Proceedings of the 40th Annual Meeting on Association for Computational Linguistics},
  pages={311--318},
  year={2002},
  organization={ACL}
}

@inproceedings{lin2004rouge,
  title={ROUGE: A Package for Automatic Evaluation of Summaries},
  author={Lin, Chin-Yew},
  booktitle={Text Summarization Branches Out: Proceedings of the ACL-04 Workshop},
  pages={74--81},
  year={2004},
  organization={ACL}
}

@misc{phi2,
  title        = {Phi-2: The surprising power of small language models},
  author       = {Microsoft Research},
  year         = {2024},
  howpublished = {\url{https://www.microsoft.com/en-us/research/blog/phi-2-the-surprising-power-of-small-language-models/}},
  note         = {Accessed: 2025-10-16}
}

@article{jiang2024mistral,
  title        = {Mistral 7B},
  author       = {Jiang, Albert Q. and Nguyen, Huu and Scao, Teven Le and others},
  journal      = {arXiv preprint arXiv:2310.06825},
  year         = {2024}
}

@misc{deepseek2024,
  title        = {DeepSeek-R1: Distilled Qwen2 1.5B for high-quality text generation},
  author       = {DeepSeek-AI Team},
  year         = {2024},
  howpublished = {\url{https://github.com/deepseek-ai}},
  note         = {Distilled variant based on Qwen2-1.5B}
}

@article{yang2024qwen2,
  title        = {Qwen2 Technical Report},
  author       = {Yang, An and Bai, Jinze and Chen, Jun and others},
  journal      = {arXiv preprint arXiv:2407.10671},
  year         = {2024}
}

@article{sur,
   author = "Couper, Mick P.",
   title = "New Developments in Survey Data Collection", 
   journal= "Annual Review of Sociology",
   year = "2017",
   volume = "43",
   number = "Volume 43, 2017",
   pages = "121-145",
   doi = "https://doi.org/10.1146/annurev-soc-060116-053613",
   url = "https://www.annualreviews.org/content/journals/10.1146/annurev-soc-060116-053613",
   publisher = "Annual Reviews",
   issn = "1545-2115",
   type = "Journal Article",
  }

@article{yang2023qwen1.5,
  title={xgen-mm (blip-3): A family of open large multimodal models},
  author={Xue, Le and Shu, Manli and Awadalla, Anas and Wang, Jun and Yan, An and Purushwalkam, Senthil and Zhou, Honglu and Prabhu, Viraj and Dai, Yutong and Ryoo, Michael S and others},
  journal={arXiv preprint arXiv:2408.08872},
  year={2024}
}

@article{touvron2023llama2,
  title        = {LLaMA 2: Open foundation and fine-tuned chat models},
  author       = {Touvron, Hugo and Martin, Louis and Stone, Kevin and others},
  journal      = {arXiv preprint arXiv:2307.09288},
  year         = {2023}
}

@misc{tinyllama2024,
  title        = {TinyLlama: Open lightweight language models pre-trained on trillion tokens},
  author       = {Zhang, Zhen and Yu, Bowen and Sun, Xu and others},
  year         = {2024},
  howpublished = {\url{https://github.com/jzhang38/TinyLlama}},
  note         = {Used as base model for PolyPersona fine-tuning}
}

@Article{vurity1,
AUTHOR = {Marasco, Emanuela and Vurity, Anudeep},
TITLE = {Late Deep Fusion of Color Spaces to Enhance Finger Photo Presentation Attack Detection in Smartphones},
JOURNAL = {Applied Sciences},
VOLUME = {12},
YEAR = {2022},
NUMBER = {22},
ARTICLE-NUMBER = {11409},
URL = {https://www.mdpi.com/2076-3417/12/22/11409},
ISSN = {2076-3417},
DOI = {10.3390/app122211409}
}

@InProceedings{adnan,
author="Adnan, Mohd
and Ahmad, Tazeem
and Rafi, Saima
and Abdullah
and Vurity, Anudeep",
editor="Nguyen, Ngoc Thanh
and Franczyk, Bogdan
and Ludwig, Andr{\'e}
and N{\'u}{\~{n}}ez, Manuel
and Treur, Jan
and Vossen, Gottfried
and Kozierkiewicz, Adrianna",
title="Multi-objective and Randomly Distributed Fuzzy Logic-Based Unequal Clustering in Heterogeneous Wireless Sensor Networks",
booktitle="Computational Collective Intelligence",
year="2024",
publisher="Springer Nature Switzerland",
address="Cham",
pages="332--345",
isbn="978-3-031-70816-9"
}

@article{faye2024k,
  title={$ k $-adaptEEGCS: Adaptive Threshold Based Automatic EEG Channel Selection},
  author={Faye, Ibrahima and Tanveer, Mohammad and Vurity, Anudeep and others},
  journal={IEEE Sensors Letters},
  year={2024},
  publisher={IEEE}
}

@inproceedings{luo2025large,
  title={Large Language Model-Empowered Adversarial Fusion for Typhoon Track Prediction},
  author={Luo, Lei and Lei, Yang and Luan, Jiahao and Vurity, Anudeep and Sriram, Sai Sumanth and Guo, Jun},
  booktitle={ICASSP 2025-2025 IEEE International Conference on Acoustics, Speech and Signal Processing (ICASSP)},
  pages={1--5},
  year={2025},
  organization={IEEE}
}

@inproceedings{vurity2024interpreting,
  title={Interpreting the fraudulence level of different finger photo presentation attack instruments},
  author={Vurity, Anudeep and Marasco, Emanuela and Ramachandra, Raghavendra and Liao, Duoduo},
  booktitle={2024 IEEE International Conference on Image Processing (ICIP)},
  pages={3236--3242},
  year={2024},
  organization={IEEE}
}

@article{abdullah2025hypergcn,
  title={HyperGCN: Interpreting Hyperscanning EEG Signals for Common Multi-Task Classification Using Graph Convolutional Networks},
  author={Abdullah, A and Faye, Ibrahima and Belhaouari, Samir Brahim and Vurity, Anudeep and Ahmad, Tazeem},
  journal={IEEE Signal Processing Letters},
  year={2025},
  publisher={IEEE}
}

@inproceedings{vurity2023new,
  title={New finger photo databases with presentation attacks and demographics},
  author={Vurity, Anudeep and Marasco, Emanuela},
  booktitle={2023 IEEE International Conference on Big Data (BigData)},
  pages={2234--2242},
  year={2023},
  organization={IEEE}
}

@inproceedings{zhang2019bertscore,
  title={BERTScore: Evaluating Text Generation with BERT},
  author={Zhang, Tianyi and Kishore, Varsha and Wu, Felix and Weinberger, Kilian Q. and Artzi, Yoav},
  booktitle={International Conference on Learning Representations (ICLR)},
  year={2020}
}

@inproceedings{li2016diversity,
  title={A Diversity-Promoting Objective Function for Neural Conversation Models},
  author={Li, Jiwei and Galley, Michel and Brockett, Chris and Gao, Jianfeng and Dolan, Bill},
  booktitle={Proceedings of NAACL-HLT},
  pages={110--119},
  year={2016},
  organization={ACL}
}

@report{finra2019,
  author       = {{FINRA Investor Education Foundation}},
  title        = {National Financial Capability Study (NFCS) 2018 Report},
  year         = {2019},
  url          = {https://www.finrafoundation.org/financial-capability}
}

@book{inglehart2014wvs,
  author       = {Inglehart, Ronald and Haerpfer, Christian and Moreno, Alejandro and Welzel, Christian and Kizilova, Kseniya and Diez-Medrano, Juan and Lagos, Marta and Norris, Pippa and Ponarin, Eduard and Puranen, Bi},
  title        = {World Values Survey: Round Six – Country-Pooled Datafile},
  publisher    = {JD Systems Institute},
  year         = {2014},
  address      = {Madrid},
  url          = {https://www.worldvaluessurvey.org/}
}

@article{chang2023survey,
  title={A Survey on Evaluation of Large Language Models},
  author={Chang, Wei and Li, Xiang and Jiang, Yi and others},
  journal={ACM Computing Surveys},
  year={2023},
  url={https://arxiv.org/abs/2307.03109}
}

@article{li2025,
  title={LLM-Generated Personas: Promise and Pitfalls for Behavioral Simulation},
  author={Li, Q. and Zhang, F. and Wang, Z.},
  journal={Journal of Artificial Intelligence Research},
  year={2025}
}

@article{millman2025,
  title={Exploring the Challenges and Potential of Synthetic Data and Survey Participants},
  author={Millman, R. and Ortega, P. and Carr, L.},
  journal={Nature Human Behaviour},
  year={2025},
  note={Forthcoming}
}

@article{shrestha2025,
  title={Beyond WEIRD: Can Synthetic Survey Participants Substitute for Humans in Global Policy Research?},
  author={Shrestha, R. and Bhatia, N. and Alam, K.},
  journal={Journal of Cross-Cultural Psychology},
  year={2025}
}

@inproceedings{jandaghi2024,
  title={Faithful persona-based conversational dataset generation with large language models},
  author={Jandaghi, Pegah and Sheng, Xianghai and Bai, Xinyi and Pujara, Jay and Sidalmed, Hakim},
  booktitle={Proceedings of the 6th Workshop on NLP for Conversational AI},
  pages={114--139},
  year={2024}
}

@inproceedings{zhang2018,
  title={Personalizing dialogue agents: I have a dog, do you have pets too?},
  author={Zhang, Saizheng and Dinan, Emily and Urbanek, Jack and Szlam, Arthur and Kiela, Douwe and Weston, Jason},
  booktitle={Proceedings of the 56th Annual Meeting of the Association for Computational Linguistics},
  volume={1},
  pages={2204--2213},
  year={2018}
}

@inproceedings{joshi2024,
  title={Persona clustering for truthful synthetic data generation},
  author={Joshi, Nitish and Wang, Xinyu and Chen, Tong and Bansal, Mohit},
  booktitle={Proceedings of the 2024 Conference of the North American Chapter of the Association for Computational Linguistics},
  pages={4567--4582},
  year={2024}
}

@inproceedings{giorgi2024,
  title={Can LLMs generate human-like beliefs? Building and evaluating synthetic belief models},
  author={Giorgi, Salvatore and Schwartz, H Andrew and Eichstaedt, Johannes C},
  booktitle={Proceedings of the 18th Conference of the European Chapter of the Association for Computational Linguistics},
  pages={2154--2175},
  year={2024}
}

@inproceedings{tjuatja2024,
  title={Do RLHF-trained LLMs believe what they say? A case study in survey response modeling},
  author={Tjuatja, Lindia and Chen, Valerie and Zou, James},
  booktitle={Proceedings of the 2024 ACM Conference on Fairness, Accountability, and Transparency},
  pages={1456--1473},
  year={2024}
}

@inproceedings{chuang2024,
  title={Belief networks: Aligning LLM agents with population-level survey statistics},
  author={Chuang, Yun-Shiuan and Goyal, Agam and Hawkins, Robert and Maki, Keith and Demszky, Dorottya},
  booktitle={Proceedings of the 62nd Annual Meeting of the Association for Computational Linguistics},
  volume={1},
  pages={10183--10207},
  year={2024}
}

@article{ge2024,
  title={PersonaHub: A billion-scale persona dataset for customized synthetic data generation},
  author={Ge, Tao and Zhang, Yeyun and Wei, Furu and Li, Zeyu and Zhang, Yuxiao and Zhang, Bo and Li, Chuanyou and Liu, Lu and Liu, Jie and Zhang, Chen and others},
  journal={arXiv preprint arXiv:2406.19213},
  year={2024}
}

\end{document}